\documentclass{article}
\usepackage{spconf,amsmath,amssymb,graphicx,booktabs,cite}
\usepackage{tikz}
\usetikzlibrary{arrows.meta,positioning,calc}
\usepackage[hidelinks]{hyperref}
\usepackage{flushend}
\usepackage{bm}
\title{Counterfactual Context Auditing for Glioma Segmentation Failure Detection Across MRI Datasets}
\name{Nguyen Linh Dan Le$^{1}$  \qquad Tran Dang Khoi$^{2}$ \qquad Nguyen Pham Hoang Le$^{3}$}
\address{$^{1}$ The University of Melbourne, Melbourne, Australia \\
$^{2}$ Industrial University of Ho Chi Minh City, Vietnam \\
$^{3}$ University of Information Technology, Vietnam National University, Vietnam}

\begin{document}
\maketitle
\begin{abstract}
Glioma segmentation models can lose accuracy across MRI datasets, while identifying the cases that require correction remains difficult. In this study, we introduce counterfactual context transport (CCT), which replaces a scan's feature statistics with those from training scans and measures the resulting changes in tumor masks. We combine this sensitivity with prediction uncertainty to rank cases for review without retraining or additional segmentation networks. Experiments on BraTS and UTSW show that context sensitivity provides evidence of segmentation failure that uncertainty in the original prediction alone does not consistently capture. Combining both signals gives the strongest average failure detection among the evaluated methods, including an ensemble of separately trained networks. These findings support using changes in tumor predictions under context replacement to help identify cases that require expert review.
\end{abstract}
\begin{keywords}
Tumor segmentation, failure detection, quality control, out of distribution, computer vision.
\end{keywords}
\vspace{-0.7em}
\section{Introduction}
MRI-based delineation of glioma informs radiotherapy planning~\cite{niyazi2023}, making it important to identify predictions that require correction. Differences in image acquisition can reduce accuracy when a model is applied to another dataset~\cite{csdg}. Reference tumor masks are usually unavailable during deployment, making these errors difficult to measure directly. Failure detection therefore identifies cases that should receive expert review~\cite{valindria2017,jaeger2023}. Existing approaches assess uncertainty in the original prediction or compare predictions from transformed inputs or different networks. Ensemble disagreement provides a strong failure-detection signal~\cite{zenk2025}, but requires several trained networks. Performance also depends on the evaluation setting and how information across voxels is combined for each scan~\cite{zenk2025}. A recent benchmark in 3D medical segmentation found that mask-level disagreement from deep ensembles provides the most reliable failure detection~\cite{zenk2025}. However, training and running multiple networks is costly, while single-model approaches based on entropy or test-time augmentation (TTA) are less reliable under distribution shift.

In this study, we investigate whether a single trained network can detect its own failures by measuring how its predictions respond to changes in acquisition context. Channel-wise feature statistics capture aspects of image style~\cite{huang2017adain,zhou2021mixstyle}. We replace these statistics in a test scan with those from training scans and measure the resulting variation in predicted masks, without changing the input image. This tests whether a prediction depends on context that may differ across datasets. Greater variation indicates sensitivity to context and provides a signal for identifying unreliable segmentations. The main contributions of our work are as follows: (1)~We introduce CCT, a training-free failure detector that matches a three-model ensemble in AUROC using one network and stored training-set feature statistics. (2)~CCT and entropy detect complementary failures; averaging their ranks gives the strongest and most consistent detection within a single-network budget. (3)~We evaluate both BraTS$\leftrightarrow$UTSW directions, using patient-level bootstrap, consistent mask-level aggregation, and multiple detection metrics, with synthetic shifts assessed separately. (4)~We show that training for invariance to the same context changes does not improve segmentation and weakens the failure-detection signal.
\begin{figure*}[t]
\centering
\includegraphics[width=\textwidth]{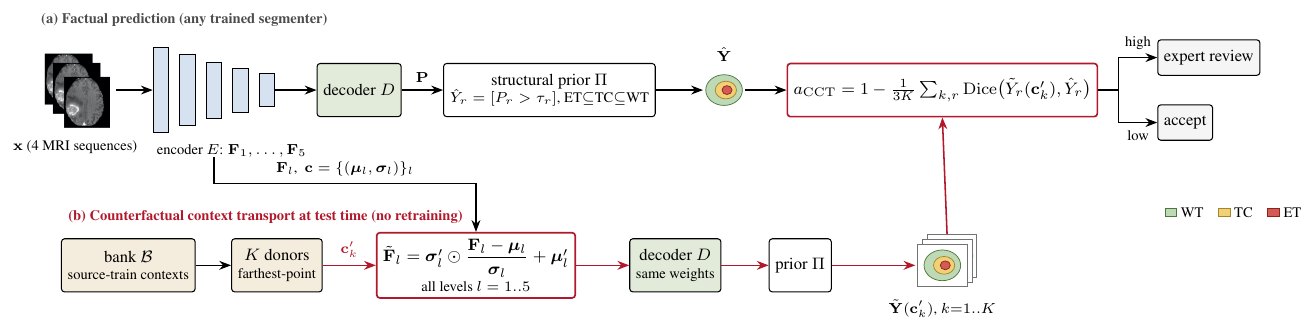}
\caption{Overview framework. CCT replaces moments at all five encoder levels with those from $K$ source training scans. The same decoder and postprocessing produce masks whose disagreement with the originals ranks cases for review.}
\label{fig:workflow}
\end{figure*}
\section{Related Work}
\noindent\textbf{Robust segmentation.} Causality-inspired and augmentation-based methods address domain shift by randomising regional appearance~\cite{csdg} or generating anatomy-preserving styles. Counterfactual segmentation has also been explored using generated healthy anatomy~\cite{cfseg}. This motivates assessing segmentation quality when a trained model is applied to new scans, however they lack the ability of direct identification of failures at deployment.

\noindent\textbf{Uncertainty and QC.} TTA~\cite{wang2019tta}, MC dropout and ensembles~\cite{lakshminarayanan2017} provide uncertainty estimates for medical segmentation, while learned QC~\cite{robinson2019} and reverse classification accuracy~\cite{valindria2017} require additional models. Zenk et~al.~\cite{zenk2025} highlighted the importance of aggregating voxel confidence into case-level scores; we use their mask-level aggregation across all detectors. AdaIN~\cite{huang2017adain} and MixStyle~\cite{zhou2021mixstyle} manipulate feature statistics for style transfer and training augmentation. We apply the same transformation at test time to a frozen segmentation network, using prediction changes to estimate case-level failure risk.

\section{Methodology}
\label{sec:method}
In our work, MRI scan $\mathbf{x}\in\mathbb{R}^{4\times|\Omega|}$ contains T1, T1ce, T2 and FLAIR, with subregion labels $\mathbf{y}\in\{0,1\}^{3\times|\Omega|}$ for necrosis (NCR), oedema (ED) and enhancing tumour (ET). The encoder $E$ produces features $\mathbf{F}_l$ at five levels (four skip levels and the bottleneck). The decoder $D$ maps these features to logits $\mathbf{s}$, giving probabilities $\mathbf{q}=\text{sig}(\mathbf{s})=(q_{\mathrm{ncr}},q_{\mathrm{ed}},q_{\mathrm{et}})$, where sig denotes sigmoid. Hats denote binary masks. Figure 1 outlines the CCT framework.

\subsection{Regions and structural prior} We obtain the BraTS regions $\mathcal{R}=\{\mathrm{WT},\mathrm{TC},\mathrm{ET}\}$ using a probabilistic OR:
\begin{equation}
\begin{aligned}
P_{\mathrm{WT}}&=1-(1-q_{\mathrm{ncr}})(1-q_{\mathrm{ed}})(1-q_{\mathrm{et}}),\\
P_{\mathrm{TC}}&=1-(1-q_{\mathrm{ncr}})(1-q_{\mathrm{et}}),\qquad P_{\mathrm{ET}}=q_{\mathrm{et}}.
\end{aligned}
\label{eq:regions}
\end{equation}
The prior $\Pi$ thresholds each region as $\hat Y_r=\mathbf{1}[P_r\ge\tau_r]$, then enforces nesting by union: $\hat Y_{\mathrm{TC}}\leftarrow\hat Y_{\mathrm{TC}}\cup\hat Y_{\mathrm{ET}}$, followed by $\hat Y_{\mathrm{WT}}\leftarrow\hat Y_{\mathrm{WT}}\cup\hat Y_{\mathrm{TC}}$. It removes ET components smaller than $k_{\min}$ voxels and repeats nesting, ensuring $\hat Y_{\mathrm{ET}}\subseteq\hat Y_{\mathrm{TC}}\subseteq\hat Y_{\mathrm{WT}}$. Removed ET components retain their support in TC and WT. We select $\tau_r$ and $k_{\min}$ per model on source validation data and fix them for all targets.

\subsection{Context as feature moments} For each level, $\bm\mu_l$ and $\bm\sigma_l$ are the channel-wise spatial mean and standard deviation of $\mathbf{F}_l$, with $\epsilon=10^{-5}$ inside the square root. We define context as $\mathbf{c}=\{(\bm\mu_l,\bm\sigma_l)\}_{l=1}^{5}$ and content as $(\mathbf{F}_l-\bm\mu_l)/\bm\sigma_l$. Linear probes recover acquisition information more accurately from context than pooled content (BraTS site balanced accuracy: 0.49 vs.\ 0.32, chance 0.17; UTSW domain: 0.76 vs.\ 0.58). Both also encode disease information (grade: 0.77 vs.\ 0.82). We assess the influence of tumour features by recomputing moments outside the predicted WT (Table 3).

\begin{table*}[t]
\centering
\footnotesize
\setlength{\tabcolsep}{5.0pt}
\begin{tabular}{lcccccccccc}
\toprule
& & \multicolumn{4}{c}{AUROC per setting}
& \multicolumn{4}{c}{Mean across settings} & \\
\cmidrule(lr){3-6}
\cmidrule(lr){7-10}
Signal & Models
& B$\to$B & B$\to$U (357) & U$\to$B & U$\to$U
& AUROC & AUPRC & AURC & Rec@20\% & Lowest\\
\midrule
Entropy & 1
& .871 {\scriptsize[.77,.95]}
& .807 {\scriptsize[.76,.84]}
& .760 {\scriptsize[.62,.88]}
& .905 {\scriptsize[.85,.96]}
& .836 & .596 & .126 & .562 & .760\\
TTA disagreement & 1
& .911 {\scriptsize[.78,.99]}
& .893 {\scriptsize[.86,.92]}
& .790 {\scriptsize[.61,.93]}
& .857 {\scriptsize[.78,.92]}
& .863 & .710 & .118 & .650 & .790\\
Ensemble disagr. & 3
& .930 {\scriptsize[.81,1.0]}
& \textbf{.899} {\scriptsize[.87,.92]}
& .808 {\scriptsize[.64,.95]}
& .852 {\scriptsize[.77,.92]}
& .872 & .703 & \textbf{.114} & .652 & .808\\
\midrule
CCT (ours) & 1
& \textbf{.932} {\scriptsize[.86,.99]}
& .893 {\scriptsize[.86,.92]}
& \textbf{.842} {\scriptsize[.71,.94]}
& .826 {\scriptsize[.73,.91]}
& .873 & .672 & .121 & .638 & \textbf{.826}\\
CCT + entropy (ours) & 1
& .923 {\scriptsize[.84,.98]}
& .895 {\scriptsize[.87,.92]}
& .826 {\scriptsize[.69,.93]}
& \textbf{.920} {\scriptsize[.86,.96]}
& \textbf{.891} & \textbf{.716} & \textbf{.114}
& \textbf{.674} & \textbf{.826}\\
\bottomrule
\end{tabular}
\caption{Comparisons.
AUROC includes 95\% patient-bootstrap intervals.
Summary columns show the mean across settings and lowest AUROC.
Rec@20\% measures failures captured within a 20\% review budget.
Disagreement scores use Eq.~\eqref{eq:audit}.}
\label{tab:audit}
\end{table*}

\subsection{Counterfactual context transport} Given donor context $\mathbf{c}'=\{(\bm\mu'_l,\bm\sigma'_l)\}$, we replace feature moments at every level using adaptive instance normalisation~\cite{huang2017adain}, then apply the unchanged decoder:
\begin{equation}
\tilde{\mathbf{F}}_l(\mathbf{c}')=\bm\sigma'_l\odot\frac{\mathbf{F}_l-\bm\mu_l}{\bm\sigma_l}+\bm\mu'_l,\quad
\tilde{\mathbf{s}}(\mathbf{c}')=D\big(\tilde{\mathbf{F}}_{1:5}(\mathbf{c}')\big).
\label{eq:cct}
\end{equation}
The context bank $\mathcal{B}$ contains source training scans only. We select a fixed donor set $\mathcal{S}\subset\mathcal{B}$ of size $K$ by farthest-point sampling in the standardised bank, starting from its most typical entry ($K=4$). Each transported prediction passes through Eq.~\eqref{eq:regions} and the same prior $\Pi$, producing $\tilde Y_r(\mathbf{c}'_k)$. We measure its disagreement with the original prediction:
\begin{equation}
a_{\mathrm{CCT}}=1-\frac{1}{K|\mathcal{R}|}\sum_{k=1}^{K}\sum_{r\in\mathcal{R}}\text{Dice}\big(\tilde Y_r(\mathbf{c}'_k),\hat Y_r\big)
\label{eq:audit}
\end{equation}
with $\text{Dice}(\emptyset,\emptyset)=1$. Higher scores indicate greater sensitivity to context. CCT requires no retraining and adds $K$ decoder passes.

\subsection{Fused audit score} We consider a budget of one trained network and one encoder pass, combining CCT with entropy from the original prediction. Let $H_2$ denote binary entropy in bits and $\Omega_{\mathrm{roi}}$ the predicted WT, falling back to the brain mask when WT is empty:
\begin{equation}
a_{\mathrm{ent}}=\frac{1}{3|\Omega_{\mathrm{roi}}|}\sum_{j\in\{\mathrm{ncr},\mathrm{ed},\mathrm{et}\}}\ \sum_{v\in\Omega_{\mathrm{roi}}}H_2\big(q_j(v)\big).
\label{eq:ent}
\end{equation}
Entropy measures uncertainty within the predicted lesion, whereas CCT measures sensitivity to context changes. We combine these complementary signals without fitting weights:
$a_{\mathrm{fused}}=\frac12(rk(a_{\mathrm{CCT}})+rk(a_{\mathrm{ent}}))$,
where $rk$ denotes normalised rank. TTA requires three additional encoder passes and the ensemble requires three networks, exceeding this budget. For evaluation, ranks are computed within each test batch. For individual-scan deployment, replacing $rk$ with the empirical CDF of a held-out source set requires neither other test scans nor labels and gives similar results (Table 2).

\vspace{-0.8em}
\section{Experiments and Results}

\subsection{Implementation details}
BraTS~2020 and UTSW-Glioma~\cite{utsw} use patient-disjoint splits of 258/37/74 and 512/37/71, stratified by grade and ET presence. UTSW validation/test labels are manually corrected; BraTS$\to$UTSW evaluates all 357 corrected cases, unseen during source training. Scans are resampled to 1mm, brain-cropped, padded, resized to $128^3$ and z-scored within the brain. MedNeXt-S~\cite{mednext} (kernel 3) is trained for 80 epochs with AdamW ($10^{-3}$), five warm-up epochs, $96^3$ patches, batch size two and mixed precision. Results use three seeds, except UTSW-trained augmentation/TRACE variants and cost experiments, which use one. Timings use a T4. Failures are the worst 20\% by mean regional Dice per setting and seed. Metrics include AUROC, AUPRC, recall at 20\% review and AURC (risk $=1-$Dice)~\cite{jaeger2023,zenk2025}. Paired patient bootstraps (2,000 resamples) preserve grouped seeds and conditions with fixed failure labels. Cross-setting means are descriptive because patients overlap. Bias field, gamma, noise and low-resolution shifts at two severities are evaluated separately on BraTS test scans.

\begin{table}[t]
\centering
\footnotesize
\setlength{\tabcolsep}{3pt}
\begin{tabular}{lcccccccc}
\toprule
& \multicolumn{2}{c}{B$\to$B}
& \multicolumn{2}{c}{B$\to$U}
& \multicolumn{2}{c}{U$\to$B}
& \multicolumn{2}{c}{U$\to$U}\\
\cmidrule(lr){2-3}
\cmidrule(lr){4-5}
\cmidrule(lr){6-7}
\cmidrule(lr){8-9}
Model & Dice & HD & Dice & HD & Dice & HD & Dice & HD\\
\midrule
MedNeXt-S
& 83.9 & 13.2
& 70.8 & 42.7
& 84.4 & 12.5
& \textbf{73.3} & \textbf{38.2}\\
+ style aug.
& \textbf{84.5} & \textbf{11.6}
& \textbf{73.4} & \textbf{39.9}
& 84.5 & \textbf{9.8}
& 72.8 & 40.9\\
+ TRACE training
& 83.7 & 12.1
& 70.9 & 41.6
& \textbf{84.8} & 10.2
& 72.0 & 41.3\\
\bottomrule
\end{tabular}
\caption{Mean Dice (\%) and HD95 (mm) over WT/TC/ET}
\label{tab:seg}
\end{table}

\begin{table}[t]
\centering
\footnotesize
\setlength{\tabcolsep}{5pt}
\begin{tabular}{lcccc}
\toprule
Variant & Models & Time (s) & ID & OOD\\
\midrule
Ensemble disagreement
& 3 & 1.06 & .921 & .838\\
TTA disagreement
& 1 & 1.25 & .922 & .780\\
\midrule
CCT, $K=1$
& 1 & \textbf{0.72} & .953 & .840\\
CCT, $K=2$
& 1 & 1.00 & \textbf{.967} & .860\\
CCT, $K=4$ (default)
& 1 & 1.51 & \textbf{.967} & .867\\
\midrule
CCT + entropy, $K=1$
& 1 & \textbf{0.72} & .955 & .885\\
CCT + entropy, $K=4$
& 1 & 1.51 & .954 & \textbf{.899}\\
\midrule
$K=4$, bottleneck only
& 1 & 1.57 & .872 & .875\\
\bottomrule
\end{tabular}
\caption{Cost for a BraTS-trained model.}
\label{tab:abl}
\end{table}

\subsection{Segmentation and transport invariance}
BraTS$\to$UTSW shows the clearest loss of segmentation accuracy
(Table~\ref{tab:seg}). Style augmentation improves this transfer,
whereas TRACE provides no consistent segmentation benefit and
reduces CCT's failure-detection performance. This suggests that
training for transport invariance can suppress a useful audit
signal without improving segmentation. 

\subsection{Failure detection}
\begin{figure}[t]
\centering
\includegraphics[width=0.94\linewidth]{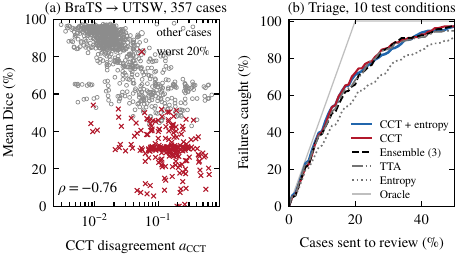}
\caption{(a) Mean Dice versus CCT disagreement on 357 UTSW
cases; $\rho$ denotes Spearman correlation.
(b) Failures captured as the review budget increases across
clean BraTS, eight synthetic shifts and UTSW.}
\label{fig:audit}
\end{figure}

CCT achieves higher mean AUROC than entropy alone
(0.873 vs 0.836) and comparable performance to the
three-model ensemble (0.872), using one trained network
(Table~\ref{tab:audit}). It outperforms entropy in both
cross-dataset directions and achieves the highest AUROC
on UTSW$\to$BraTS, demonstrating the value of context
sensitivity for detecting failures under dataset shift.
Combining CCT with entropy further improves detection,
giving the highest mean AUROC (0.891), AUPRC and recall
within the review budget, while matching the ensemble's
mean AURC. Fusion also retains entropy's strong performance
on UTSW$\to$UTSW, providing more consistent detection
across settings. On BraTS$\to$UTSW, the largest evaluation,
CCT alone shows no clear paired AUROC advantage over the
ensemble and has lower AUPRC and worse AURC.
Source-reference scoring performs similarly to test-batch
ranking (0.889 and 0.832 on B$\to$U and
U$\to$B, against 0.895 and 0.826), supporting individual-scan scoring.
Figure~\ref{fig:audit} shows that greater CCT disagreement
is associated with poorer segmentation, supporting
prioritisation for expert review. Synthetic shifts provide
further evidence, although these perturbations do not
constitute independent patient cohort.

\subsection{Computational cost and design}
With one donor, fusion achieves higher AUROC on both evaluated sets at lower inference cost than the ensemble (0.72 vs.\ 1.06\,s; Table~\ref{tab:abl}). The default $K=4$ is slower but requires only one trained network. Fusion is less sensitive to donor selection and bottleneck-only transport, while tumour-excluded moments retain useful detection. Adding TTA offers limited gains at greater cost. Independent validation remains necessary because fusion was selected on this benchmark.\section{Conclusion and Future Work}
CCT audits a single trained segmenter without retraining by measuring prediction sensitivity to feature-statistic changes. Combined with entropy, it achieves the strongest mean failure detection across the evaluated settings, supporting prioritisation of cases for expert review. Its main limitations are that feature statistics mix imaging and disease information, and the donor bank restricts the contexts examined. Future work will investigate acquisition-specific representations, broader donor coverage and validation on independent cohorts.

\bibliographystyle{IEEEbib}
\bibliography{main}
\end{document}